\documentclass[preprint,12pt]{elsarticle}

\usepackage[dvipsnames]{xcolor}
\usepackage{tikz}
\usepackage{amsmath}
\usepackage{amsthm}
\usepackage{pgfplots}
\usepackage{pgfplotstable}
\usetikzlibrary{pgfplots.groupplots}
\pgfplotsset{compat=1.15}
\usetikzlibrary{arrows.meta}
\pgfplotsset{compat=newest}
\usepackage{dirtytalk}
\usepackage{nicefrac} 
\usepackage{algorithm}
\usepackage{colortbl}
\usepackage{algpseudocodex}
\usepackage{multirow}
\usepackage[normalem]{ulem}
\usepackage{booktabs}

\definecolor{C0}{HTML}{1f77b4}
\definecolor{C1}{HTML}{ff7f0e}

\newcommand{\withdagger}[1]{#1\(^{\dagger}\)}

\newtheorem{remark}{Remark} 
\usepackage{graphicx} 
\usepackage{subcaption} 
\usepackage{amssymb}
\usepackage[pagebackref, breaklinks, colorlinks=true, citecolor=MidnightBlue, linkcolor=BrickRed, urlcolor=NavyBlue]{hyperref}
\usepackage[capitalize]{cleveref}
\Crefname{table}{Tab.}{Tabs.}
\Crefname{section}{Sec.}{Secs.}
\usepackage{physics}

\journal{Pattern Recognition}

\begin{document}

\begin{frontmatter}



\title{PASS: Peer-Agreement based Sample Selection for Training with Noisy Labels}


\author[label1]{Arpit Garg}
\author[label2]{Cuong Nguyen}
\author[label1]{Rafael Felix}
\author[label3]{Thanh-Toan Do}
\author[label2]{Gustavo Carneiro}

\affiliation[label1]{organization={Australian Institute for Machine Learning, University of Adelaide},
            country={Australia}}

\affiliation[label2]{organization={Centre for Vision, Speech and Signal Processing, University of Surrey},
            country={United Kingdom}}
\affiliation[label3]{organization={Department of Data Science and AI, Monash University},
            country={Australia}}

\begin{abstract}

The prevalence of noisy-label samples poses a significant challenge in deep learning, inducing overfitting effects. This has, therefore, motivated the emergence of learning with noisy-label (LNL) techniques that focus on separating noisy- and clean-label samples to apply different learning strategies to each group of samples. Current methodologies often rely on the small-loss hypothesis or feature-based selection to separate noisy- and clean-label samples, yet our empirical observations reveal their limitations, especially for labels with instance dependent noise (IDN). An important characteristic of IDN is the difficulty to distinguish the clean-label samples that lie near the decision boundary (i.e., the \emph{hard} samples) from the noisy-label samples. We, therefore, propose a new noisy-label detection method, termed \emph{Peer-Agreement based Sample Selection} (PASS), to address this problem. Utilising a trio of classifiers, PASS employs consensus-driven peer-based agreement of two models to select the samples to train the remaining model. PASS is easily integrated into existing LNL models, enabling the improvement of the detection accuracy of noisy- and clean-label samples, which increases the classification accuracy across various LNL benchmarks.
\footnote{The code will be open-sourced upon the acceptance of the paper.}

\end{abstract}



\end{frontmatter}


\section{Introduction}
\label{sec:introduction}
In deep neural networks (DNNs) and machine learning, it is commonly recognised that having an adequate amount of labelled training data and computational resources leads to exceptional outcomes in various fields~\cite{goodfellow2016deep}, such as computer vision, natural language processing, and in the medical domain. 
However, such positive outcomes have been achieved predominantly through the utilisation of meticulously curated datasets that possess labels of exceptional quality. The collection of such high-quality labels, particularly for large datasets, can be exorbitantly costly in real-world scenarios~\cite{song2022learning}. Therefore, cheaper alternative labelling methods, including crowd-sourcing~\cite{song2022learning} and meta-data mining~\cite{feng2021ssr}, have gained traction, but they result in substandard labelling~\cite{song2022learning}. Although these techniques reduce costs and expedite labelling, they are susceptible to data mislabelling~\cite{song2022learning}.

Erroneous labels can potentially degrade the performance of DNNs by inducing overfitting through the phenomenon of memorisation~\cite{li2020dividemix, cordeiro2023longremix}. This issue has led to the development of innovative learning algorithms to tackle the problem of noisy-labelling. 
Within the domain of noisy-labels, many methods have emerged~\cite{li2020dividemix, Garg_2023_WACV}, each tailored to tackle the challenges posed by distinct noise settings, namely instance-independent noise (IIN)~\cite{li2020dividemix} and instance-dependent noise (IDN)~\cite{xia2020part}. Early studies of noisy-labels operated under the assumption that the label noise was IIN, where mislabelling occurred regardless of the information on the visual classes present in images~\cite{li2020dividemix}. Conventional IIN methods often employ a transition matrix which comprises a predetermined probability of flipping between pairs of labels~\cite{xia2020part}. However, recent studies have progressively redirected the field's attention towards the more realistic scenario of IDN~\cite{ Garg_2023_WACV}, where label noise depends on both clean-label and the image information.

Previous techniques for mitigating the impact of noisy-label samples frequently involve manually selecting clean samples to form a clean validation set~\cite{ren2018learning}. The difficulty in obtaining clean validation samples, particularly for problems with many classes, has motivated recent studies to leverage semi-supervised learning methods without relying on clean validation sets~\cite{li2020dividemix, sachdeva2023scanmix}. Other approaches incorporate robust loss functions~\cite{song2022learning}, designed specifically to operate effectively with clean or noisy-labels, as well as probabilistic modeling approaches that model the data generation process~\cite{Garg_2023_WACV}. Furthermore, training regularisation~\cite{liu2020early} imposes a penalty term on the loss function during training, thus reducing overfitting and ameliorating generalisation. Various techniques integrate sample selection strategies as a key algorithmic step~\cite{Garg_2023_WACV, li2020dividemix}, allowing the detection of clean and noisy-label samples. 
A widespread criterion for this sample selection process is the loss value between the prediction of the trained classifier and its label, by which it is generally assumed that the noisy-label data exhibits a large loss~\cite{li2020dividemix,kim2021fine} or a higher magnitude of the gradient during training~\cite{song2022learning}.

Furthermore, feature-based sample selection techniques relying on the similarity to the principal components of feature representations~\cite{kim2021fine} or K nearest neighbor (KNN) classification in the feature space~\cite{feng2021ssr} have also been considered for the sample selection criteria.
However, we empirically show in \cref{fig:empirical_graph} that the separation of clean, but difficult-to-classify samples from noisy-label samples remains a challenge for these sample selection processes~\cite{wei2020combating}, particularly for problems with high noise rates. 
The use of peer classifiers for noisy-label learning problems has been investigated to avoid confirmation bias~\cite{han2018co, li2020dividemix}, but not to select clean and noisy-label samples. 
\begin{figure*}[t]
    \centering
    \includegraphics[width=\linewidth]{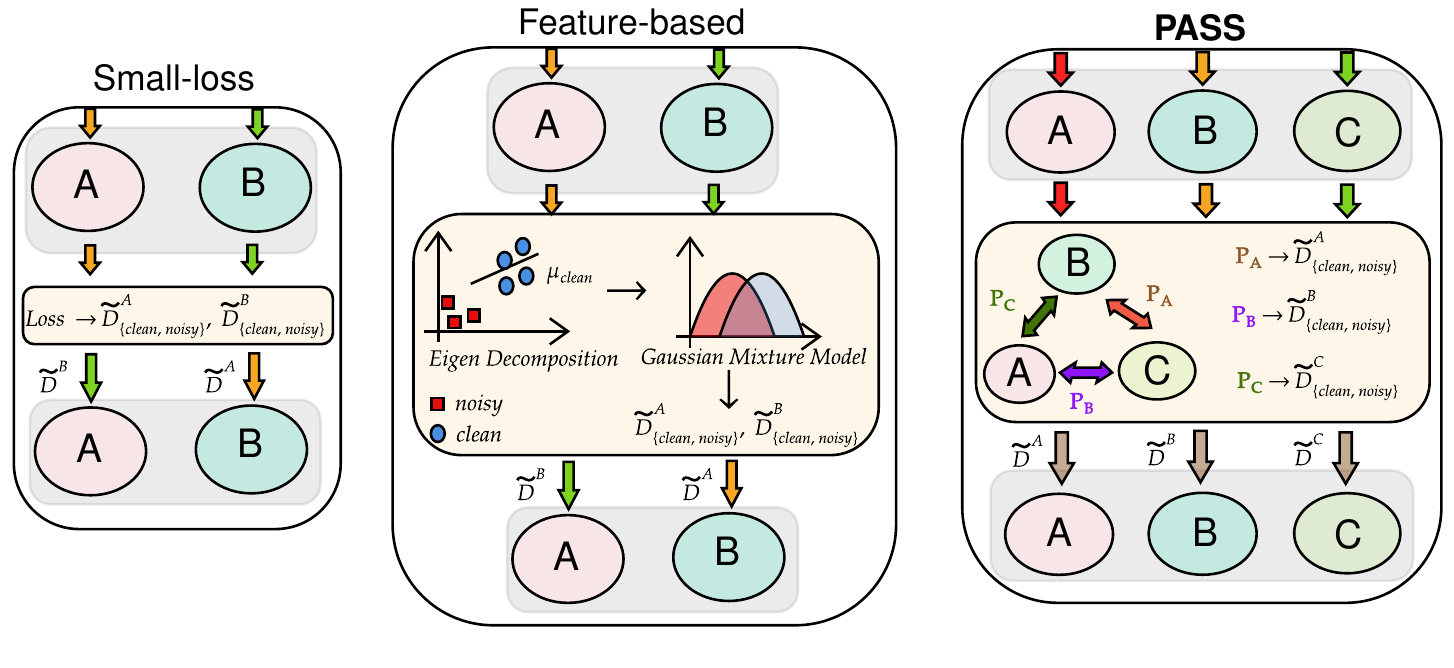}
    \caption{A visual comparison between various sample selection strategies, including: \emph{(left)} the small-loss approaches~\cite{li2020dividemix} typically have two networks, where one model uses the loss values of samples to select clean and noisy instances to train the other and vice versa; \emph{(middle)} feature-based approaches~\cite{kim2021fine} detect outliers of features belonging to one class to detect noisy-label samples; \emph{(right)} PASS consists of three networks where samples agreed by two networks are considered as clean and used for training the third network.}   
    \label{fig:motivation}
\end{figure*}
We argue in this paper that the prediction agreement between peer classifiers is more effective in selecting clean and noisy-label samples than previous approaches, because, intuitively, such an agreement is unlikely to happen, except when the classifiers agree on the clean-label. 

In this paper, we propose a new sample selection criterion based on the predictive probability agreement between peer classifiers.
In our proposed method, we train three classifiers simultaneously using the agreement between two classifiers to select samples to train the remaining classifier, as shown in~\cref{fig:motivation}.
This sample selection is based on a thresholding algorithm~\cite{otsu1979threshold} that distinguishes samples based on the degree of agreement between the peer classification predictions. Our proposed method, named as \emph{peer-agreement-based sample selection} (PASS), can easily be integrated into existing models in noisy-label learning, such as InstanceGM~\cite{Garg_2023_WACV}, DivideMix~\cite{li2020dividemix}, SSR~\cite{feng2021ssr}, FaMUS~\cite{xu2021faster}, AugDesc~\cite{nishi2021augmentation}, and Contrast-to-Divide (C2D)~\cite{zheltonozhskii2022contrast}. Our primary contributions can be delineated as follows:

\begin{itemize}
\item we propose a new noisy-label sample selection method, PASS, that differentiates clean and noisy-label samples through prediction agreement between peer classifiers, and
\item we demonstrate that our method can be easily adapted to existing models, including InstanceGM~\cite{Garg_2023_WACV}, DivideMix~\cite{li2020dividemix}, SSR~\cite{feng2021ssr}, FaMUS~\cite{xu2021faster}, AugDesc~\cite{nishi2021augmentation}, and C2D~\cite{zheltonozhskii2022contrast}, where we show that PASS enhances the performance of various SOTA approaches on various benchmarks, comprising both simulated and real-world datasets, such as CIFAR-100~\cite{krizhevsky2009learning}, CIFAR-N~\cite{wei2022learning}, Animal-10N~\cite{song2019selfie}, Red mini-ImageNet from Controlled Noisy Web Labels (CNWL)~\cite{xu2021faster}, Clothing1M~\cite{xiao2015learning}, mini-WebVision~\cite{li2017webvision}, and ImageNet~\cite{deng2009imagenet}.
\end{itemize}

It is imperative to clarify that our proposition does not involve the introduction of a new learning with noisy-label algorithm. Instead, we suggest a new method for selecting noisy-label training samples to substantially improve the efficacy of preexisting LNL algorithms, as shown in our experimental section. The empirical evidence supporting this selection mechanism is delineated 
and elaborated in \cref{subsec:empirical_subsec_cifar} and \cref{subsec:empirical_subsec_clothing}. Upon completion of the selection phase, our approach adheres to a robust training algorithm specifically designed to handle noisy-labels, which facilitates the development of a more reliable model.

\section{Related Work}
\label{sec:related_work}
\paragraph{\textbf{Learning with noisy-label (LNL)}} DNNs have been demonstrated to be highly effective in fitting randomly-labelled training data, which may result in overfitting~\cite{cordeiro2023longremix}. Consequently, when presented with clean-label testing data, these networks exhibit poor generalisation~\cite{li2020dividemix}. To address this challenge, numerous studies have explored supervised learning in a label noise setting, leading to the development of various techniques, such as robust loss functions~\cite{xiao2015learning}, sample selection~\cite{li2020dividemix}, robust regularisation~\cite{xiao2015learning}, and robust architectures~\cite{song2022learning}. Our work primarily aims at addressing the issue of noisy-label learning for DNNs, with a particular emphasis on sample selection methods. We provide a brief review of the methods proposed in the literature that fall into this category. To maximise the utility of the entire dataset during training, including the noisy-label subset, recent methods have incorporated semisupervised training~\cite{li2020dividemix, Garg_2023_WACV} techniques, such as MixMatch~\cite{berthelot2019mixmatch}. This involves treating the clean subset as labelled and the noisy subset as unlabelled. However, to accomplish this, a sample selection stage is required to classify samples as clean or noisy. Although this approach is well-motivated and often effective, it is vulnerable to cumulative errors from the selection process, particularly when there are numerous unreliable samples in the training set~\cite{song2022learning}. Therefore, sample selection methods often employ multiple clean-label sample classifiers to enhance their robustness against such cumulative errors~\cite{kim2021fine, feng2021ssr}. In most cases, distinguishing between hard-to-classify clean and noisy-label samples poses a significant challenge. This issue is highlighted by recent research~\cite{song2022learning, kim2021fine}, which notes that model performance can significantly deteriorate due to sampling errors~\cite{kim2021fine}.

\paragraph{\textbf{Sample selection}} The sample selection process falls into two categories, namely loss-based sampling~\cite{jiang2018mentornet} and feature-based sampling~\cite{feng2021ssr}. Loss-based sample selection~\cite{li2020dividemix, zheltonozhskii2022contrast} involves the application of the small-loss trick, which hypothesises that noisy data tend to incur a high loss due to the model's difficulty in correctly classifying such data. However, these methods require the adjustment of a small-loss threshold to enable the selection of training samples. Such a challenge motivated the development of complex strategies, such as the filtration of samples using a small-loss over a number of training epochs~\cite{li2020dividemix}.  
In feature-based selection~\cite{kim2021fine, feng2021ssr}, clean and noisy samples are classified using features extracted from the input data. For example, clean samples can be identified by KNN \cite{feng2021ssr, wu2020topological}, or distance to eigenvectors~\cite{kim2021fine}. Others follow sampling-based technique that employs an adversarial filtering-based approach to eliminate spurious artifacts in a dataset~\cite{le2020adversarial}. Furthermore, resampling procedures which learns a weight distribution to favor difficult instances for a given feature representation~\cite{li2019repair} is suggested to reduce representation bias. However, these two techniques rely solely on theoretical claims or require meticulous fine-tuning of complex hyperparameters to accommodate the type and magnitude of the noise present, leading to significant performance degradation when incorrect selections are made~\cite{song2022learning}.

\paragraph{\textbf{Data reliability}}
The reliability of data is crucial to develop a computational model or to support an empirical claim~\cite{he2022towards}. Over the years, research has focused on various ways to select reliable samples from unreliable datasets based on peer effects and social networks~\cite{newman2018network}. 
In an unreliable data environment, peer-based sample selection has emerged as a promising approach to train models and select high-confidence samples~\cite{song2022learning}. This method involves using a group of models to collectively judge and select samples to train the model~\cite{malach2017decoupling}. 
This technique aims to improve the performance of the model in scenarios that involve unreliable data, while also mitigating the influence of confirmation bias~\cite{song2022learning}. 
Peer-based sample selection has the potential to enhance the accuracy and reliability of learning systems in situations where data quality is uncertain~\cite{ramesh2022peer}. 
When different models produce consistent results, it indicates that they have a similar understanding of categories and can be expected to perform consistently~\cite{ramesh2022peer}. 
However, it is imperative to consider that attaining agreement does not invariably guarantee validity; nevertheless, it is probable that they would agree on reliable samples to a greater extent~\cite{ramesh2022peer}. 
Although our work draws inspiration from the aforementioned approaches, our fundamental aim is to address the issue of noisy-label image classification through the application of a novel sample selection.

\section{Methodology}\label{sec:methodology}
\begin{figure*}[t]
    \centering
    \includegraphics[width=\linewidth]{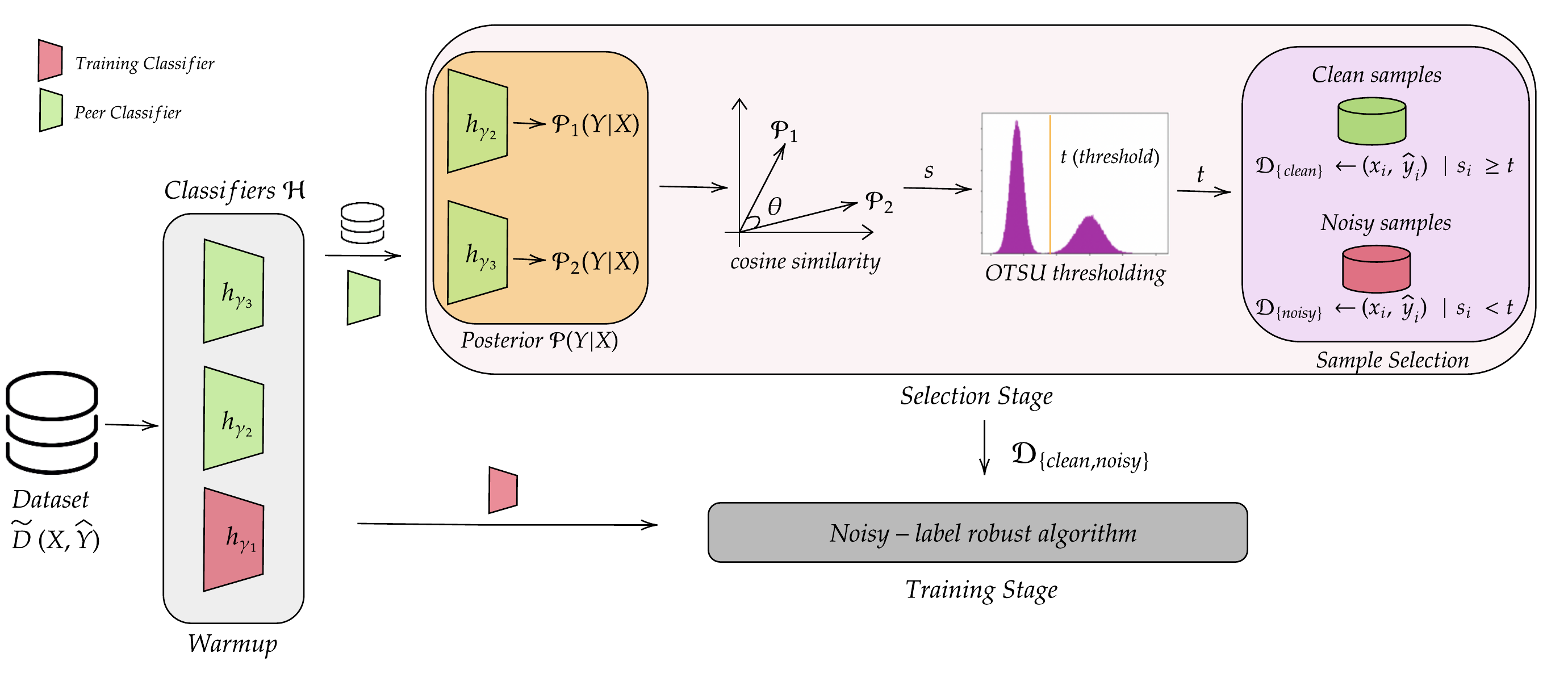}
    \caption{Our proposed method, PASS, consists of three classifiers trained in a round-robin fashion, with two classifiers (\(h_{\gamma_{2},\gamma_{3}}\) in green) being used to select samples for training the other classifier (\(h_{\gamma_{1}}\) in red).
    The training process begins with a warm-up of all classifiers, followed by the sample selection stage. During the selection stage, the peer classifiers calculate the prediction agreement using cosine similarity between their posterior distributions, followed by Otsu's thresholding~\cite{otsu1979threshold} to automatically find the threshold $t$ to select the clean set ${D}_{\text{clean}}$ and noisy set ${D}_{\text{noisy}}$. In the training stage, we follow the robust noisy-label training algorithm.}
    \label{fig:architecture}
\end{figure*}
\subsection{Problem definition}\label{problem_definition}
Formally, we define the instance space as \(\mathcal{X}\) and their respective label space as \(\mathcal{Y}\). The training set is represented by $\Tilde{{D}} = \{(\mathbf{x}_{i}, \hat{\mathbf{y}}_{i}) \}^{n}_{i=1}$, where $\mathbf{x}_{i} \in \mathcal{X} \subseteq \mathbb{R}^{d}$ represents an instance, and $\hat{\mathbf{y}}_{i} \in \mathcal{Y} = \{ \hat{\mathbf{y}}: \hat{\mathbf{y}} \in \{0,1\}^{C} \wedge \pmb{1}^{\top} 
\hat{\mathbf{y}} = 1 \}$, denotes the $C$-dimensional one-hot vector representation of the corresponding noisy-label. 
In the conventional classification problem, $\Tilde{{D}}$ is used to train a classifier $h_{\gamma}: {X} \to \Delta_{C-1}$, parameterised by $\gamma \in \Gamma \subseteq \mathbb{R}^{\abs{\gamma}}$ with $\Delta_{C - 1}$ representing the \((C - 1)\)-dimensional probability simplex. In noisy-label learning, noisy-label data $\Tilde{{D}}$ are exploited to
obtain a model $h_{\gamma}$ that can accurately predict the clean-label of samples in a test set. 
\begin{algorithm}[t!]
    \caption{Sample Selection and Training of Noise-Robust Classifiers}\label{procedure:Proposed_Procedure}
    \begin{algorithmic}[1]
    { 
        \Procedure{PASS}{$\Tilde{D}, \Psi, E$}
            \LComment{$\Tilde{D}=\{(\mathbf{x}_i, \hat{\mathbf{y}}_i)\}_{i=1}^{n}$: noisy-labelled dataset}
            \LComment{$E$: total number of epochs}
            \LComment{$\Psi$: training algorithm to use, e.g., DivideMix}
            \State Initialise three classifiers randomly: \(\{\gamma_{j}\}_{j = 1}^{3}\)
            \For{$e = 1:E$} \Comment{for each epoch}
                \LComment{Select clean/noisy samples for classifier $h_{\gamma_{1}}$}
                \State $L_{1}, U_{1} \gets$ \Call{Sample-Selection}{$ \Tilde{D}, h_{\gamma_{2}}, h_{\gamma_{3}}$} \Comment{$L$: clean, $U$: noisy}
                \State $\mathsf{L}_{(1)}^{\mathrm{(\Psi)}} \gets$ \Call{Loss}{$ L_{1}, U_{1}, h_{\gamma_{1}}$}
                \State \(\gamma_{1} \gets\) \Call{SGD}{$\mathsf{L}_{(1)}^{\mathrm{(\Psi)}}, \gamma_{1}$} \Comment{update parameter}
                \State Repeat for $h_{\gamma_{2}}$ and $h_{\gamma_{3}}$
            \EndFor
            \State \Return ${\gamma_{1}}, {\gamma_{2}}, {\gamma_{3}}$
        \EndProcedure
        \Statex  
        \Function{Sample-Selection}{$\Tilde{D}, h_{\gamma_{j}},h_{\gamma_{k}}$}
            \State $\mathbf{s} \gets \pmb{0}$  \Comment{vector to store cosine similarity}
            \For{each $ (\mathbf{x}_{i}, \hat{\mathbf{y}}_{i})$ in $\Tilde{D}$}
                \State $\mathbf{s}_{i} \gets$ \Call{Cosine-Similarity}{$h_{\gamma_{j}}(\mathbf{x}_{i}), h_{\gamma_{k}}(\mathbf{x}_{i})$}
            \EndFor
            \State $t \gets$ \Call{Otsu}{$\mathbf{s}$}  \Comment{Find thresholding with Otsu}
            \State $D_{\text{clean}} \gets \varnothing, D_{\text{noisy}} \gets \varnothing$
            \For{ each $(\mathbf{x}_{i}, \hat{\mathbf{y}}_{i})$ in $\Tilde{D}$}
                \If{$\mathbf{s}_i \geq t$}  \Comment{Highly-agreed}
                    \State $D_{\mathrm{clean}} \gets D_{\text{clean}} \cup {(\mathbf{x}_i, \hat{\mathbf{y}}_i)}$
                \Else  \Comment{Lowly-agreed}
                    \State $D_{\mathrm{noisy}} \gets D_{\mathrm{noisy}} \cup {(\mathbf{x}_i, \hat{\mathbf{y}}_i)}$
                \EndIf
            \EndFor
            \State \Return $D_{\mathrm{clean}}, D_{\mathrm{noisy}}$
        \EndFunction
        }
    \end{algorithmic}
\end{algorithm}

\subsection{Reliability based sample selection}
To enhance the lucidity of our explanation, we begin by delineating our methodology for sample selection. As shown in~\cref{fig:architecture}, our proposed method, PASS, requires at least three classifiers: \(\{h_{\gamma_{k}}\}_{k = 1}^{3}\), to select reliable samples via peer agreement. In particular, all classifiers consistently rotate between the roles of peers and training classifiers. It is also important to note that we have randomly initialised the classifiers to reduce the chances of confirmation bias~\cite{li2020dividemix}. Another important note is that our sample selection approach can be easily integrated into various models in LNL, as we demonstrate in \cref{sec:experiments}.

The output of the $k$-th classifier, denoted by \(h_{\gamma_{k}} (\mathbf{x}_{i})\), represents the probability of $\mathbf{y}_{i}$ given $\mathbf{x}_{i}$. The predictive probability agreement between two peer classifiers: \(h_{\gamma_{l}} \) and \(h_{\gamma_{m}} \), on a data point \(\mathbf{x}_{i}\) is defined as the cosine similarity of the two predictions made by the two moels: 
\begin{equation}\label{eq:cosine_sim}
    \mathbf{s}_{i} = \operatorname{agreement} \left( h_{\gamma_{l}}, h_{\gamma_{m}} \vert \mathbf{x}_{i} \right) = \frac{h_{\gamma_{l}}(\mathbf{x}_{i})^{\top}  \, h_{\gamma_{m}}(\mathbf{x}_{i})}{\norm{h_{\gamma_{l}}(\mathbf{x}_{i}) } \, \norm{h_{\gamma_{m}}(\mathbf{x}_{i}) }},
\end{equation}
where \(\norm{.}\) denotes the Frobenius norm.

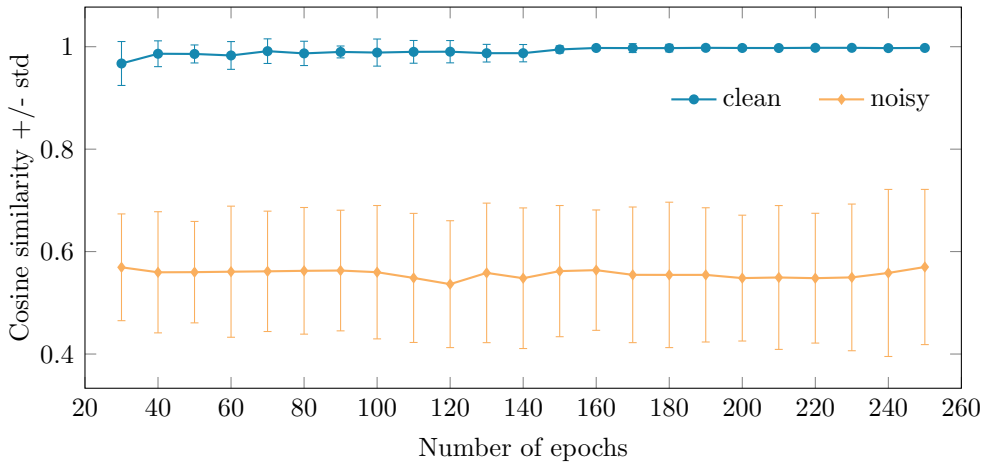
\begin{figure}[t!]
    \centering
    \begin{tikzpicture}
        \pgfplotstableread[col sep=comma, header=true]{cosine_data.txt} \myTable
        \begin{axis}[
            height=0.37\linewidth,
            width=0.85\linewidth,
            xlabel={Number of epochs},
            xlabel style={font=\footnotesize},
            xticklabel style={font=\footnotesize},
            ylabel={Cosine similarity +/- std},
            ylabel style={font=\footnotesize, yshift=-0.5em},
            yticklabel style={font=\footnotesize},
            xmin=20,
            xmax=260,
            legend style={draw=none, font=\footnotesize, yshift=-2em, /tikz/every even column/.append style={column sep=1em}},
            legend columns=2,
            scale only axis
        ]
            \addlegendentry{clean};
            \addplot[mark=*, mark size=1.5pt, mark repeat=0, mark phase=0, MidnightBlue!70, thick, solid] plot [error bars/.cd, y dir=both, y explicit] table[x={epochs}, y={cleanCosine}, y error={cleanStd}]{\myTable};

            \addlegendentry{noisy};
            \addplot[mark=diamond*, mark size=1.5pt, BurntOrange!70, thick, solid] plot [error bars/.cd, y dir=both, y explicit] table[x={epochs}, y={noisyCosine}, y error={noisyStd}]{\myTable};
            
            
        \end{axis}
    \end{tikzpicture}
    \caption{Probability agreement of clean- and noisy-label samples on CIFAR100~\cite{krizhevsky2009learning} at \(0.5\) IDN~\cite{xia2020part} using DivideMix-PASS. It shows the cosine similarity with standard deviation between clean samples (\textcolor{MidnightBlue}{blue}) and the cosine similarity with standard deviation between noisy samples (\textcolor{BurntOrange}{orange}) over the number of epochs. }
    \label{fig:cosinehypothesis}
\end{figure}

\begin{remark}
\label{thm:remark}
    Noisy-label samples will likely have small predictive probability agreement values in \cref{eq:cosine_sim}, as empirically shown in~\cref{fig:cosinehypothesis}. The reason is that the different influences of noisy-label samples on predictive probabilities can result in disparate outcomes between peer classifiers (see~\cref{fig:cosinehypothesis} - orange). Moreover, clean data is likely to have high predictive probability agreement (see~\cref{fig:cosinehypothesis} - blue). According to many studies in peer classifier agreement~\cite{ramesh2022peer}, it is recommended to select clean data based on a high peer classifier agreement, but noisy data should be selected cautiously based on low peer classifier agreement.
\end{remark}

Leveraging the remark above, we partition the training set \(\Tilde{{D}}\) into a clean set \(\Tilde{{D}}_{\text{clean}}\) and a noisy set  \(\Tilde{{D}}_{\text{noisy}}\) based on the value of the cosine similarity in \cref{eq:cosine_sim}. This partition can be achieved by any thresholding algorithm, with the clean set comprising data points exhibiting high cosine similarity values and the noisy set comprising data points exhibiting low cosine similarity values. In our case, we use the global thresholding technique known as Otsu's algorithm \cite{otsu1979threshold}. Compared to other clustering and thresholding algorithms (K-Means and the Gaussian Mixture Model (GMM)), Otsu's algorithm \cite{otsu1979threshold} is advantageous, as it can find the optimal clustering, and hence, provides PASS with the best performance, as shown in the ablation studies under \cref{sec:ablation}. This thresholding algorithm automatically estimates an optimal threshold $t$ to divide the data samples into two classes, namely \emph{clean} ($\mathbf{s}_{i} \geq t$ or most likely agreed) and \emph{noisy} data ($\mathbf{s}_{i} < t$ or most unlikely agreed). Further detailed explanation of  Otsu's algorithm \cite{otsu1979threshold} can be found in~\cref{sec:appendix_otsu}. 

Once clean and noisy samples have been selected, we employ  noisy-label learning training algorithms from the literature (\cite{Garg_2023_WACV,li2020dividemix,feng2021ssr,xu2021faster,nishi2021augmentation,zheltonozhskii2022contrast}). These algorithms are employed to test the efficacy of our proposed sampling approach in experiments. Our training procedure is succinctly described in~\cref{procedure:Proposed_Procedure} and visually portrayed in~\cref{fig:architecture}.

\subsection{Otsu's algorithm}\label{sec:appendix_otsu}

Otsu's algorithm~\cite{otsu1979threshold} aims to estimate the threshold that partitions data samples by maximising the between-class variance and minimising the within-class variance. The Otsu's thresholding stands out as a notably straight-forward and advantageous global thresholding approach. The Otsu's formula for finding the optimal threshold \(t^{*}\) is the following: 
\begin{equation}\label{eq:Otsu}
    t^* = \arg\max_{t}( \sigma_B^2(t) ),
\end{equation}
where $t$ is the threshold value, $\sigma_B^2(t)$ is the between-class variance for threshold $t$, computed as 
$$\sigma_B^2(t) = w_1(t)w_2(t)(\mu_1(t) - \mu_2(t))^2,$$
with $w_1(t)$ and $w_2(t)$ representing the weights of the clean and noisy classes (calculated as fractions of the data on each side of the threshold), and $\mu_1(t),\mu_2(t)$ representing the mean values of cosine similarity in the clean and noisy classes, respectively. The effectiveness of the Otsu's clustering is compared other clustering approaches in \cref{sec:ablation}.

\section{Experiments} \label{sec:experiments}

This section presents an extensive experimental evaluation of PASS. We present the datasets, implementation details, and results over several LNL benchmarks, followed by an ablation study.

\begin{table}[t!]
    \centering
    \caption{Test accuracy (\%) on CIFAR-100~\cite{krizhevsky2009learning} subject to various IDN noise rates~\cite{xia2020part}. The results were obtained from~\cite{Garg_2023_WACV}, wherein the base model (\(*\)) results are denoted in \textit{italics}. \(\dagger\) represents the SOTA and \textbf{PASS} represents our approach with mentioned baselines.}
    \label{tab:cifar}
    \begin{tabular}{l c c c c c}
        \toprule
        \textbf{Method} & \textbf{0.20} & \textbf{0.30} & \textbf{0.40} & \textbf{0.45} & \textbf{0.50} \\
        \midrule
        CE~\cite{yao2021instance}  & 30.42 & 24.15 & 21.45 & 15.23 & 14.42\\
        USDNL~\cite{xu2023usdnl} & 64.82 & 61.35 & 55.82 & - & 46.00 \\
        PTD-R-V~\cite{xia2020part}  & 65.33 & 64.56\ & 59.73 & - & 56.80 \\
        MentorNet~\cite{jiang2018mentornet}  & 38.91 & 34.23 & 31.89 & 27.53 & 24.15\\
        \midrule
        DivideMix*~\cite{li2020dividemix}  & \textit{77.07} & \textit{76.33} & \textit{70.80} & \textit{57.78}  & \textit{58.61}\\
        \rowcolor{Gray!25} \textbf{DivideMix-PASS}  & \textbf{77.41} & \textbf{76.58} & \textbf{75.07} & \textbf{72.91}  & \textbf{72.27}\\
        \midrule
        InstanceGM*~\cite{Garg_2023_WACV}  & \textit{79.69} & \textit{79.21} & \textit{78.47} & \textit{77.49}  & \textit{77.19}\\
        \rowcolor{Gray!25} \textbf{InstanceGM-PASS} & \withdagger{\textbf{81.02}} & \withdagger{\textbf{80.33}} & \withdagger{\textbf{79.28}} & \withdagger{\textbf{78.69}} & \withdagger{\textbf{78.26}}\\
        \bottomrule
    \end{tabular}
\end{table} 

\subsection{Datasets}\label{subsec:dataset}

The experiments are performed on many common datasets in LNL, including CIFAR-100~\cite{krizhevsky2009learning}, CIFAR-N~\cite{wei2022learning}, Animal-10N~\cite{song2019selfie}, Red mini-ImageNet~\cite{jiang2020beyond}, Clothing-1M~\cite{xiao2015learning} and mini-WebVision~\cite{li2020dividemix}.
\paragraph{CIFAR-100} The dataset consists of \(50,000\) training images and \(10,000\) testing images with each image having a size of \(32 \times 32 \times 3\) pixels, distributed evenly into 100 categories. This dataset does not possess label noise by default, so we follow the \emph{part-dependent label noise} setting~\cite{xia2020part} to simulate various IDN noise rates: \(\{0.2, 0.3, 0.4, 0.45, 0.5\}.\)

\paragraph{CIFAR-10N and CIFAR-100N} The datasets are created by relabelling both the original CIFAR-10 and CIFAR-100~\cite{wei2022learning} datasets using the Amazon Mechanical Turk (M-Turk) labelling service.
The CIFAR-10N dataset includes five distinct noise rate options, from which we have selected the \say{\emph{worst}} version (noise rate of \(40.21\)\%). In the CIFAR-100N dataset, we considered \say{\emph{fine}} labels with an overall noise level of \(40.20\)\%.

\paragraph{Animal-10N} This is a real-world dataset including \(10\) animal categories, with \(5\) pairs of animals sharing similar appearances, such as \emph{chimpanzee} and \emph{orangutan}. The dataset has an estimated label noise rate of \(8\%\), and it comprises of \(50,000\) training images and \(10,000\) test images. In the experiments, we do not perform data augmentation to be consistent with the standard setup~\cite{song2019selfie} for a fair evaluation.
 
\paragraph{Red mini-ImageNet} The dataset is a subset of the real-world CNWL dataset, which is mainly established to examine the impact of label noise rates on image classification. This dataset includes 100 categories where each categories consists of \(600\) colour images. To ensure an equitable comparison to previous studies, all images have been resized to 32\(\times\)32 pixel\({}^{2}\). There are various noise rates ranging from \(0\% \text{ to } 80\%\). We focused on the noise rates of \(40\%, 60\%, \text{ and } 80\%\) to maintain consistency with the existing literature~\cite{Garg_2023_WACV, xu2021faster}.
\paragraph{Clothing1M} This is also a real-world dataset consisting of 1 million training images collected from \(14\) distinct online shopping website categories. There is an estimated \(38.5\%\) noise level in this dataset's labels, which are derived from the surrounding text. To ensure comparability, we used downsized images to \(256 \times 256\) pixel\(^{2}\), as per the prevalent format in previous works~\cite{Garg_2023_WACV, li2020dividemix}. There are \(50,000, 14,000, \text{ and } 10,000\) manually authenticated training, validation, and testing samples, respectively. We excluded clean training and validation sets during training. We only use the clean test set for evaluation, following the literature~\cite{Garg_2023_WACV, li2020dividemix}.

\begin{table}[t!]
    \centering
    \caption{Test accuracy (\%) on CIFAR-N~\cite{wei2022learning}, where results of other models are from~\cite{wei2022learning}. The \textbf{PASS} base model  is DivideMix~\cite{li2020dividemix} (\(*\) with results in \textit{italics}) and \(\dagger\) represents the SOTA.}
    \label{table:cifar100N}
    \begin{tabular}{l c c}
        \toprule
        \bfseries Method & \bfseries CIFAR10N-W & \bfseries CIFAR100N-F  \\
        \midrule
        CE~\cite{liu2022robust} & 77.69 & 55.50\\
        CAL~\cite{zhu2021clusterability} & 85.36 & 61.73\\
        ELR~\cite{liu2020early} & 91.09 & 66.72\\
        \midrule
        DivideMix*~\cite{li2020dividemix} & \textit{92.56} & \textit{71.13}\\
        \rowcolor{Gray!25} \textbf{DivideMix-PASS}  & \withdagger{\textbf{94.02}} & \withdagger{\textbf{72.03}} \\
        \bottomrule
    \end{tabular}
\end{table}
\begin{table}[t]
        \centering
        \caption{Test accuracy (\%) of various approaches on Animal-10N~\cite{song2019selfie} with baseline (\(*\) and outcomes in \textit{italics}). The other results are  from~\cite{feng2021ssr}. \textbf{PASS} represents our approach with baseline DivideMix~\cite{li2020dividemix} and SSR~\cite{feng2021ssr}, and \(\dagger\) denotes the SOTA.}
        \label{table:Animal10N}
        \begin{tabular}{l c}
            \toprule
            \bfseries Method & \bfseries Test Accuracy (\%) \\
            \midrule
            CE~\cite{zhang2021learning} & 79.4 \\
            SELFIE~\cite{song2019selfie} & 81.8 \\
            PLC~\cite{zhang2021learning} & 83.4 \\
            Jigsaw-ViT~\cite{chen2023jigsaw} & \textbf{89.0}\\
            \midrule
            DivideMix*~\cite{li2020dividemix} & \textit{81.40}\\
            \rowcolor{Gray!25}\textbf{DivideMix-PASS} & \textbf{82.90} \\
            \midrule
            SSR*~\cite{feng2021ssr} & \textit{88.5}\\
            \rowcolor{Gray!25} \textbf{SSR-PASS} & \withdagger{\textbf{89.2}} \\
            \bottomrule
        \end{tabular}
    \end{table}
\paragraph{Mini-WebVision} The dataset consists of \(65,944\) colour images taken from the initial \(50\) categories of the WebVision dataset~\cite{li2017webvision}, with images reduced to \(256 \times 256\) pixels. In the experiments, we follow the standard benchmark by evaluating on the clean validation sets of both mini-WebVision and the equivalent \(50\) categories from the ImageNet dataset~\cite{deng2009imagenet}.

\subsection{Implementation}\label{subsec:implementation}

All methods are implemented in the PyTorch framework and executed on the NVIDIA RTX 3090 GPU computing platform. Baseline models are selected based on their accuracy and compatibility with the dataset under consideration. For CIFAR-100, the InstanceGM~\cite{Garg_2023_WACV} and DivideMix~\cite{li2020dividemix} models are used because both have demonstrated to be highly accurate. For CIFAR-N, the DivideMix~\cite{li2020dividemix} model is used. For Animal-10N, SSR~\cite{feng2021ssr} is selected as the base model. For Red mini-ImageNet, a hybrid approach using FaMUS~\cite{xu2021faster} with two evaluation versions, one with and one without DINO self-supervision~\cite{caron2021emerging} is employed. For Clothing-1M, AugDesc~\cite{nishi2021augmentation} model is used. For mini-WebVision, C2D~\cite{zheltonozhskii2022contrast} is employed as the base model. Unless otherwise stated, default hyperparameters and network architectures are as specified in their corresponding papers.

\subsection{Comparisons on Benchmarks}\label{subsec:baseline}
In this section, we perform a comparison study on IDN benchmarks and real-world noisy-label benchmarks.

\subsubsection{IDN Benchmark}\label{subsubsec:idn_benchmark}
 
In \cref{tab:cifar}, a comparative analysis is presented showcasing the performance of the proposed method, PASS, against various SOTA techniques on the CIFAR-100 IDN benchmark~\cite{xia2020part}.
In particular, PASS outperforms these models by approximately between \(1.2\%\) to \(14\%\) at \(0.50\) noise rate. 

\subsubsection{Real-world noisy-label benchmarks}\label{subsubsec:real-world}
In~\cref{table:cifar100N,table:Animal10N,table:RedMini,table:clothing1M,table:miniWebvision}, we showcase the results of our proposed method on CIFAR-N~\cite{wei2022learning}, Animal-10N~\cite{song2019selfie}, Red mini-ImageNet~\cite{jiang2020beyond}, Clothing1M~\cite{xiao2015learning}, mini-WebVision~\cite{li2020dividemix} and ImageNet~\cite{krizhevsky2012imagenet}. Overall, PASS demonstrates superior performance or competitiveness with current SOTA models.
The results also show that PASS exhibits a high degree of flexibility and can be easily integrated into existing LNL models.

\begin{table}[t!]
    \centering
    \caption{Test accuracy (\%) on Red mini-ImageNet (CNWL)~\cite{jiang2020beyond}. The additional results of the model are from~\cite{Garg_2023_WACV}. We show \textbf{PASS} (ours) using DivideMix~\cite{li2020dividemix} and FaMUS~\cite{xu2021faster} ($\ast$ and the results in \textit{italics}, \(\dagger\) represents the SOTA) without and with self-supervision (SS)~\cite{caron2021emerging}.}
    \label{table:RedMini}
    \begin{tabular}{l c c c}
        \toprule
        \multirow{2}{*}{\textbf{Red mini-ImageNet}} & \multicolumn{3}{c}{\textbf{Noise rate}} \\
        \cmidrule{2-4}
         & \textbf{0.4} & \textbf{0.6} & \textbf{0.8} \\
        \midrule
        CE~\cite{xu2021faster} &  42.70 & 37.30 & 29.76 \\
        MentorMix~\cite{jiang2020beyond} &  47.14 & 43.80 & 33.46 \\
        InstanceGM~\cite{Garg_2023_WACV} &  52.24 & 47.96 & 39.62 \\
        \midrule
        DivideMix~\cite{li2020dividemix} & \textit{46.72} & \textit{43.14} & \textit{34.50} \\
        \rowcolor{Gray!25}\textbf{DivideMix-PASS}  & \textbf{53.02} & \textbf{48.01} & \textbf{38.62} \\
        \midrule
        FaMUS$\ast$~\cite{xu2021faster} & \textit{51.42} &  \textit{45.10} & \textit{35.50} \\
        \rowcolor{Gray!25}\textbf{FaMUS-PASS}  &  \withdagger{\textbf{53.40}} & \withdagger{\textbf{48.04}} & \withdagger{\textbf{40.08}} \\
        \midrule
        \multicolumn{4}{l}{\textbf{With self-supervised learning}} \\
        \midrule
        InstanceGM-SS*~\cite{Garg_2023_WACV} & 56.37 & 53.21 & 44.03 \\
        \rowcolor{Gray!25}\textbf{FaMUS-SS-PASS}  & \withdagger{\textbf{56.48}} & \withdagger{\textbf{53.53}} & \withdagger{\textbf{44.32}} \\
        \bottomrule
    \end{tabular}
\end{table}

\begin{table}[t!]
    \centering
    \caption{Test accuracy (\%) of competing strategies on Clothing1M~\cite{xiao2015learning}. In the experiments, only noisy-labels are used for training. The base models used are DivideMix~\cite{li2020dividemix}, AugDesc~\cite{nishi2021augmentation} and FINE~\cite{kim2021fine} with results in \textit{italics}. \textbf{PASS} results are within 1\%, and \(\dagger\) represents the SOTA.}
    \label{table:clothing1M}
    \begin{tabular}{l c}
        \toprule
        \textbf{Clothing1M} & \textbf{Test Accuracy (\%)} \\
        \midrule
        Nested-CoTeaching~\cite{chen2021boosting} & 74.90 \\
        MLC~\cite{zheng2021meta} & \withdagger{75.78} \\
        \midrule
        DivideMix~\cite{li2020dividemix} & \textit{74.76}\\
        \rowcolor{Gray!25}\textbf{DivideMix-PASS}  & \textbf{74.82}\\
        \midrule
        FINE~\cite{kim2021fine} & \textit{74.37} \\
        \rowcolor{Gray!25}\textbf{FINE-PASS} & \textbf{74.42} \\
        \midrule
        AugDesc-WAW*~\cite{nishi2021augmentation} & \textit{74.72} \\
        \rowcolor{Gray!25}\textbf{AugDesc-WAW-PASS}  & \textbf{74.81} \\
        AugDesc-SAW*~\cite{nishi2021augmentation} & \textit{75.11} \\
        \rowcolor{Gray!25}\textbf{AugDesc-SAW-PASS}  & \textbf{75.13} \\
        \bottomrule
    \end{tabular}
\end{table}


\begin{table}[t!]
    \centering
    \caption{Test accuracy (\%) on mini-WebVision~\cite{li2020dividemix} and validation on ImageNet~\cite{deng2009imagenet}. Base models are DivideMix~\cite{li2020dividemix} and Contrast-to-Divide(C2D)~\cite{zheltonozhskii2022contrast} represented by \(*\) with results in \textit{italics}, whilst \textbf{C2D-PASS} is our proposed results, and \(\dagger\) represents the SOTA.
    }\label{table:miniWebvision}
    \begin{tabular}{l c c c c }
        \toprule
        \multirow{2}{*}{\bfseries Dataset} & \multicolumn{2}{c}{\bfseries Mini-WebVision} & \multicolumn{2}{c}{\bfseries ImageNet} \\ 
        \cmidrule(lr){2-3} \cmidrule(lr){4-5}
        & \textbf{Top-1} & \textbf{Top-5} & \textbf{Top-1} & \textbf{Top-5} \\
        \midrule
        BtR~\citep{smart2023bootstrapping} & 80.88 & 92.76 & 75.96 & 92.20 \\
        SSR~\cite{feng2021ssr} & \withdagger{80.92} & 92.80 & 75.76 & 91.76\\
        \midrule
        DivideMix~\cite{li2020dividemix} & \textit{77.32} & \textit{91.64} & \textit{75.20} & \textit{91.64} \\
        \rowcolor{Gray!25}\textbf{DivideMix-PASS} & \textbf{78.64} & \textbf{92.20} & \textbf{75.91} & \textbf{91.80} \\
        \midrule
        C2D*~\cite{zheltonozhskii2022contrast} & \textit{79.42} & \textit{92.32} & \textit{78.57} & \textit{93.04}\\
        \rowcolor{Gray!25} \textbf{C2D-PASS}  & \textbf{80.72} & \withdagger{\textbf{92.91}} & \withdagger{\textbf{79.32}} & \withdagger{\textbf{93.20}}  \\
        \bottomrule
    \end{tabular}
\end{table} 

In more detail, \cref{table:cifar100N,table:Animal10N} present the results obtained by PASS with their corresponding baselines in CIFAR-N~\cite{wei2022learning} and Animal-10N~\cite{song2019selfie}, respectively. 
It is noteworthy that the results from PASS are shown to improve all baselines, exhibiting competitive performance across both datasets. 

\cref{table:RedMini} reports the results on Red mini-ImageNet~\cite{xu2021faster} using our PASS method with baseline model FaMUS~\cite{xu2021faster} in two different setups: 1) without pretraining (upper section of the table), and 2) with self-supervised (SS) pre-training (lower section of the table). SS pre-training relies on DINO~\cite{caron2021emerging} using the unlabelled Red mini-ImageNet dataset to ensure a fair comparison with InstanceGM~\cite{Garg_2023_WACV}. The results demonstrate that PASS can effectively improve performance and achieve SOTA outcomes on Red mini-ImageNet~\cite{xu2021faster}.

\begin{table}[t!]
    \centering
    \caption{This ablation study shows the test accuracy \(\%\) on CIFAR-100~\cite{krizhevsky2009learning} under IDN~\cite{xia2020part} at noise rate of \(0.5\). We show the result of our method using various clustering algorithms (Gaussian Mixture Model (GMM), K-Means, and  Otsu's thresholding~\cite{otsu1979threshold}) under the DivideMix~\cite{li2020dividemix} baseline.}
    \label{tab:ablation_cifar}
    \begin{tabular}{l c}
        \toprule
        \bfseries DivideMix-PASS &  \bfseries Test Accuracy (\%) \\
        \midrule
        GMM & 64.10 \\
        K-Means & 66.21 \\
        \rowcolor{Gray!25}\textbf{OTSU} & \textbf{72.27} \\
        \bottomrule
    \end{tabular}
\end{table}

\cref{table:clothing1M} shows the result on Clothing-1M, where two different training setups named AugDesc-WAW~\cite{nishi2021augmentation} and AugDesc-SAW~\cite{nishi2021augmentation} are used. PASS is found to be easily adaptable to both versions, delivering highly competitive results compared to the existing methods.

Furthermore, \cref{table:miniWebvision} presents the results obtained by PASS on mini-WebVision~\cite{li2017webvision} and ImageNet~\cite{krizhevsky2012imagenet}. In particular, the results are shown to improve all baselines and exhibit competitive performance across the entire dataset. 

\section{Empirical Analysis}
\subsection{Ablation Study on Clustering Algorithms}\label{sec:ablation}
This section present our ablation study on different algorithms that cluster the peer agreement in \cref{eq:cosine_sim} to partition the training dataset into a clean and a noisy subsets. The ablation study is conducted on CIFAR-100~\cite{krizhevsky2009learning} in IDN settings~\cite{xia2020part} with a noise rate of \(0.5\). Two other clustering algorithms, namely K-Means and GMM, are considered in this study with results shown in \cref{tab:ablation_cifar}. Overall, the performance of K-Means and GMM are lower than Otsu's algorithm, which could be attributed to their nature: K-Means and GMM are optimally local (depending on initialisation and stopping criteria), while Otsu's algorithm is a global one due to its exhaustive search. It is worth noting that using GMM and K-Means offers improvements of approximately \(2\%\) accuracy w.r.t. the baseline method, DivideMix~\cite{li2020dividemix}. However, using these clustering techniques can still restrict the classification accuracy since Otsu's thresholding~\cite{otsu1979threshold} enables a further improvement in accuracy of approximately \(6\%\).

\subsection{Computational Time}
\label{sec:computational_time}
We show a training time comparison between various base models~\cite{li2020dividemix, Garg_2023_WACV, feng2021ssr, xu2021faster, nishi2021augmentation, zheltonozhskii2022contrast} and their PASS variants in \cref{table:computation}. Overall, PASS has an overhead due to the usage of multiple classifiers compared to the corresponding baselines. However, this aspect of PASS is mitigated by its satisfactory performance in terms of running time, particularly when executed using half-precision, which stands favorably against its baselines.

\begin{table}[t!]
    \centering
    \caption{Training time (in hours) of the base models and base models with \textbf{PASS} (ours) on various datasets}
    \label{table:computation}
    \begin{tabular}{l l r r}
        \toprule
        \textbf{Models} & \textbf{Dataset} & \textbf{Base} & \textbf{PASS} \\
        \midrule
        DivideMix~\cite{li2020dividemix} & CIFAR-100  & 7.5 & 9.8 \\
        InstanceGM~\cite{yao2021instance} & CIFAR-100 & 31.2 & 34.0\\
        SSR~\cite{feng2021ssr} & Animal-10N & 6.5 & 9.8\\
        FaMUS~\cite{xu2021faster} & Red mini-ImageNet & 12.0 & 14.2 \\
        FINE~\cite{kim2021fine} & Clothing1M & 30.3 & 34.1\\ 
        AugDesc~\cite{nishi2021augmentation} & Clothing1M & 29.6 & 30.1\\
        FINE~\cite{kim2021fine} & Mini-WebVision & 41.5 & 44.7\\
        C2D~\cite{zheltonozhskii2022contrast} & Mini-WebVision & 42.2 & 44.1\\
        \bottomrule
    \end{tabular}
\end{table}

\subsection{Statistical Hypothesis Testing on Models' Performances}
We perform a statistical hypothesis testing to determine if the integration of PASS into other SOTA methods is effective. Our study compares three models (i.e., DivideMix~\cite{li2020dividemix}, InstanceGM~\cite{Garg_2023_WACV}, and DivideMix-PASS (ours)) on ten datasets (i.e., CIFAR-100~\cite{krizhevsky2009learning} with noise rates of \(0.2, 0.3, 0.4, 0.45, \text{ and } 0.5\), Red mini-ImageNet~\cite{jiang2020beyond} at \(0.4, 0.6, \text{ and } 0.8\) noise rates, Clothing 1M~\cite{xiao2015learning}, and Animal-10N~\cite{song2019selfie}) using one metric, standard accuracy. Generally, a hypothesis testing consists of:
\begin{itemize}
    \item a \emph{null} hypothesis denoting that all means of models' performance are equal, and
    \item an \emph{alternative} hypothesis denoting that at least one of the models performs differently.
\end{itemize}
The conclusion of such a hypothesis testing, of course, holds statistically under a certain significant level (usually 0.05).

One straight approach to compare the performance of several models on many datasets is ANOVA (analysis of variance). However, ANOVA assumes that data follows a normal distribution, which might not hold in our case. Hence, we employ the Friedman test -- a non-parametric hypothesis testing -- as an alternative one.


The Friedman test with a significance level of \(0.1\), yielded a test statistic of \(16.20\) and a p-value of \(3.035 \times 10^{-4}\), leading us to reject the null hypothesis that all methods perform equally well. This suggests that at least one of the methods significantly differs from the others in terms of performance. 
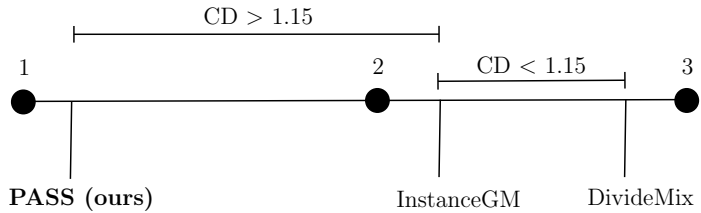
\begin{figure}[t]
    \centering
    \scalebox{0.75}{
    \tikzset{every picture/.style={line width=0.75pt}} 

\begin{tikzpicture}[x=0.75pt,y=0.75pt,yscale=-1,xscale=1]

\draw    (27.54,233.52) -- (463.44,232.34) ;
\draw  [fill={rgb, 255:red, 0; green, 0; blue, 0 }  ,fill opacity=1 ] (13.82,233.91) .. controls (13.82,229.54) and (17.33,226) .. (21.66,226) .. controls (25.99,226) and (29.5,229.54) .. (29.5,233.91) .. controls (29.5,238.28) and (25.99,241.82) .. (21.66,241.82) .. controls (17.33,241.82) and (13.82,238.28) .. (13.82,233.91) -- cycle ;
\draw    (54.16,233.52) -- (53.33,285.68) ;
\draw    (302.06,232.34) -- (301.22,284.49) ;
\draw    (426.83,232.34) -- (426,284.49) ;
\draw    (55.69,188.37) -- (301.21,188.29) ;
\draw [shift={(301.21,188.29)}, rotate = 179.98] [color={rgb, 255:red, 0; green, 0; blue, 0 }  ][line width=0.75]    (0,5.59) -- (0,-5.59)   ;
\draw [shift={(55.69,188.37)}, rotate = 179.98] [color={rgb, 255:red, 0; green, 0; blue, 0 }  ][line width=0.75]    (0,5.59) -- (0,-5.59)   ;
\draw  [fill={rgb, 255:red, 0; green, 0; blue, 0 }  ,fill opacity=1 ] (251.94,232.91) .. controls (251.94,228.54) and (255.45,225) .. (259.78,225) .. controls (264.12,225) and (267.63,228.54) .. (267.63,232.91) .. controls (267.63,237.28) and (264.12,240.82) .. (259.78,240.82) .. controls (255.45,240.82) and (251.94,237.28) .. (251.94,232.91) -- cycle ;
\draw  [fill={rgb, 255:red, 0; green, 0; blue, 0 }  ,fill opacity=1 ] (459.94,232.91) .. controls (459.94,228.54) and (463.45,225) .. (467.78,225) .. controls (472.12,225) and (475.63,228.54) .. (475.63,232.91) .. controls (475.63,237.28) and (472.12,240.82) .. (467.78,240.82) .. controls (463.45,240.82) and (459.94,237.28) .. (459.94,232.91) -- cycle ;
\draw    (300.5,220) -- (426.21,219.29) ;
\draw [shift={(426.21,219.29)}, rotate = 179.67] [color={rgb, 255:red, 0; green, 0; blue, 0 }  ][line width=0.75]    (0,5.59) -- (0,-5.59)   ;
\draw [shift={(300.5,220)}, rotate = 179.67] [color={rgb, 255:red, 0; green, 0; blue, 0 }  ][line width=0.75]    (0,5.59) -- (0,-5.59)   ;

\draw (16.87,204.09) node [anchor=north west][inner sep=0.75pt]   [align=left] {1};
\draw (10.5,289.62) node [anchor=north west][inner sep=0.75pt]   [align=left] {\textbf{PASS (ours)}};
\draw (271.05,291.99) node [anchor=north west][inner sep=0.75pt]   [align=left] {InstanceGM};
\draw (399.48,290.81) node [anchor=north west][inner sep=0.75pt]   [align=left] {DivideMix};
\draw (141.34,168.86) node [anchor=north west][inner sep=0.75pt]   [align=left] {CD $>$ 1.15};
\draw (255,203.09) node [anchor=north west][inner sep=0.75pt]   [align=left] {2};
\draw (463,203.09) node [anchor=north west][inner sep=0.75pt]   [align=left] {3};
\draw (324.91,202.02) node [anchor=north west][inner sep=0.75pt]   [align=left] {CD $<$ 1.15};

\end{tikzpicture}}
    \caption{Critical Difference (CD) diagram comparing DivideMix~\cite{li2020dividemix}, InstanceGM~\cite{Garg_2023_WACV}, and DivideMix-PASS (ours). Average ranks are derived from performance across datasets, with lines indicating the range of non-significant differences per the Nemenyi test. This test estimated the CD value of \(1.15\), which is used to estimate if two models are different with the significance level of \(0.1\). DivideMix-PASS demonstrates statistically significant superiority, InstanceGM~\cite{Garg_2023_WACV} and DivideMix~\cite{li2020dividemix} are not significantly different.}
    
    
    \label{fig:stats_cd_diagram}
\end{figure}

To further understand these differences, we applied the post-hoc Nemenyi test. The test results showed significant differences between some of the methods. The Critical Difference (CD) value was calculated to be approximately \(1.15\). Based on this value, the methods whose average ranks differ by at least this CD value are considered significantly different at the \(0.1\) confidence level. Our analysis indicates that PASS is significantly different from both DivideMix and InstanceGM, as denoted by the Nemenyi test p-values (\(0.001\) against both). However, there is no significant difference between DivideMix and InstanceGM, as their comparison yields a p-value of \(0.109268\), which is above our threshold for significance. This comprehensive statistical analysis illustrates (\cref{fig:stats_cd_diagram}) the comparative effectiveness of these methods in handling various types and degrees of noise in datasets, affirming that DivideMix-PASS (ours) exhibits a statistically significant improvement over the other methods under study.

\subsection{Empirical Analysis on Sample Selection}\label{subsec:sample_selection}

\begin{figure*}[t]
    \centering
    \hspace{-1em}
    \begin{subfigure}[b]{0.3\textwidth}
        \centering
        \begin{tikzpicture}
            \pgfplotstableread[col sep=comma, header=true]{csv/f1.tex} \myTable

            \begin{axis}[
                height = 0.6\linewidth,
                width = 0.8\linewidth,
                xlabel={\textnumero~of epochs},
                xlabel style={font=\scriptsize},
                xticklabel style = {font=\scriptsize},
                xmin=40,
                xmax=310,
                ylabel={F1-score},
                ylabel style={font=\scriptsize, yshift=-0.5em},
                yticklabel style = {font=\scriptsize},
                scale only axis
            ]
                \addplot[mark=none, MidnightBlue, thick, solid] table[x={epochs}, y={FINE}]{\myTable};
                \addplot[mark=none, BurntOrange, thick, densely dashed] table[x={epochs}, y={SmallLoss}]{\myTable};
                \addplot[mark=none, BrickRed, solid, thick, dashdotted] table[x={epochs}, y={PASS}]{\myTable};
            \end{axis}
        \end{tikzpicture}
        \caption{F1 Score}
        \label{fig:f1}
    \end{subfigure}
    \hfill
    \begin{subfigure}[b]{0.3\textwidth}
        \centering
        \begin{tikzpicture}
            \pgfplotstableread[col sep=comma, header=true]{csv/precision.tex} \myTable

            \begin{axis}[
                height = 0.6\linewidth,
                width = 0.8\linewidth,
                xlabel={\textnumero~of epochs},
                xlabel style={font=\scriptsize},
                xticklabel style = {font=\scriptsize},
                xmin=40,
                xmax=310,
                ylabel={Precision},
                ylabel style={font=\scriptsize, yshift=-0.5em},
                yticklabel style = {font=\scriptsize},
                legend entries={FINE, Small loss, PASS},
                legend style={draw=none, font=\tiny, /tikz/every even column/.append style={column sep=0.1em}},
                legend image post style={scale=.75},
                legend columns=2,
                legend cell align={left},
                legend style={at={(0.0,0.6)},anchor=west},
                scale only axis
            ]
                \addplot[mark=none, MidnightBlue, thick, solid] table[x={epochs}, y={FINE}]{\myTable};
                \addplot[mark=none, BurntOrange, thick, densely dashed] table[x={epochs}, y={SmallLoss}]{\myTable};
                \addplot[mark=none, BrickRed, solid, thick, dashdotted] table[x={epochs}, y={PASS}]{\myTable};
            \end{axis}
        \end{tikzpicture}
        \caption{Precision}
        \label{fig:precision}
    \end{subfigure}
    \hfill
    \begin{subfigure}[b]{0.35\textwidth}
        \centering
        \begin{tikzpicture}
            \pgfplotstableread[col sep=comma, header=true]{csv/clean.tex} \myTable

            \begin{axis}[
                height = 0.514\linewidth,
                width = 0.686\linewidth,
                xlabel={\textnumero~of epochs},
                xlabel style={font=\scriptsize},
                xticklabel style = {font=\scriptsize},
                xmin=40,
                xmax=310,
                ylabel={Ratio of clean data},
                ylabel style={font=\scriptsize, yshift=-0.5em},
                yticklabel style = {font=\scriptsize},
                scale only axis
            ]
                \addplot[mark=none, MidnightBlue, thick, solid] table[x={epochs}, y={FINE}]{\myTable};
                \addplot[mark=none, BurntOrange, thick, densely dashed] table[x={epochs}, y={SmallLoss}]{\myTable};
                \addplot[mark=none, BrickRed, solid, thick, dashdotted] table[x={epochs}, y={PASS}]{\myTable};

                \addplot[mark=none, style={densely dashed}] coordinates {
                    (50, 0.5)
                    (300, 0.5)
                };
                \node[above, color=Black, rotate=0, ] at (110, 0.5) {\scriptsize{ideal ratio}};
            \end{axis}
        \end{tikzpicture}
        \caption{Ratio of data classified as clean}
        \label{fig:ratio}
    \end{subfigure}
    \caption{Comparative analysis of the effectiveness of selecting clean or noisy samples, with reference to three metrics: (a) F1-score, (b) precision, and (c) ratio of data classified as clean. The comparison is made between PASS (dash-dot red), the small-loss approach~\cite{jiang2018mentornet} (dashed yellow), and feature-based selection FINE~\cite{kim2021fine} (solid blue) (all are based on DivideMix backbone~\cite{li2020dividemix}), implemented on the CIFAR-100 dataset\cite{krizhevsky2009learning} at \(0.5\) IDN noise rate, as described in~\cite{xia2020part}. }
    \label{fig:empirical_graph}
\end{figure*}
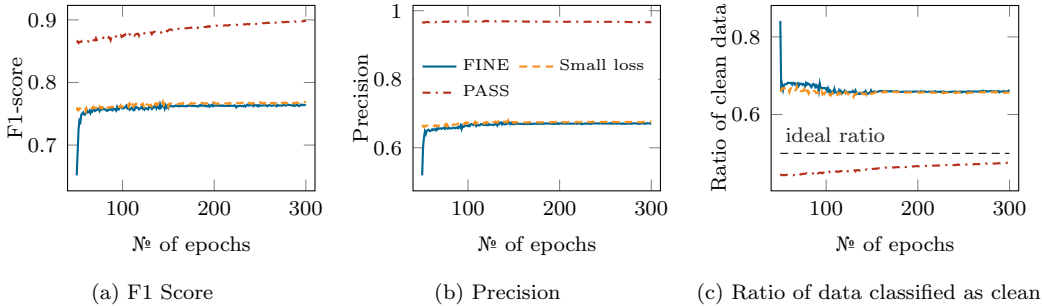

In this section, we conduct an empirical analysis of the PASS algorithm. Detailed in \cref{subsec:empirical_subsec_cifar}, the analysis compares the effectiveness of PASS against small loss~\cite{li2020dividemix} and FINE~\cite{kim2021fine} approaches on the CIFAR100~\cite{krizhevsky2009learning}, specifically at \(0.5, 0.4, \text{ and } 0.2\) IDN~\cite{xia2020part}. The comparison focuses on metrics F1 score, precision, and the ratio of clean samples, employing DivideMix~\cite{li2020dividemix} as the base model for PASS. Similarly,~\cref{subsec:empirical_subsec_clothing} extends this analysis to the Clothing1M~\cite{xiao2015learning}, a real-world dataset. Here, the comparison is between AugDesc~\cite{nishi2021augmentation}, and PASS with baseline AugDesc~\cite{nishi2021augmentation}.

\subsubsection{Analysis of PASS Performance at Various Noise Rates on CIFAR100}\label{subsec:empirical_subsec_cifar}
\paragraph{IDN setting at high noise rate (e.g., 50\%)} To empirically analyse PASS, we focus on the challenging IDN synthetic noise at \(50\%\) noise rate~\cite{xia2020part} on CIFAR-100~\cite{krizhevsky2009learning}.
\cref{fig:empirical_graph} shows three graphs to measure the performance of the clean sample classification, namely: \cref{fig:f1} shows \emph{F1 score}, \cref{fig:precision} shows \emph{precision}, and \cref{fig:ratio} shows the \emph{ratio of the data classified as clean}.
We use these graphs to compare our PASS against the small-loss hypothesis~\cite{li2020dividemix}, and feature-based approach~\cite{kim2021fine} (all using DivideMix~\cite{li2020dividemix} as the robust noisy-label training algorithm). We have only considered the methods of sample selection and have not incorporated the methods that involve sample relabeling within this analysis~\cite{feng2021ssr}. 

\cref{fig:ratio} shows the proportion of data classified as clean (by the model). 
It is evident that the small-loss~\cite{li2020dividemix} hypothesis and the feature-based approach~\cite{kim2021fine} consistently yield a ratio of around $0.70 - 0.65$ during the training process, while our approach maintains a ratio of around $0.50-0.45$. 
As we know from the setup, the optimal rate (ideal ratio) should be $\approx$ $0.50$. This indicates that our approach is more capable of identifying the correct proportion of noisy-label samples for the IDN at \(50\%\) on CIFAR-100. 
However, that proportion alone does not ensure that clean samples are accurately selected. 
Therefore, we also calculated the F1 score (\cref{fig:f1}) and the precision (\cref{fig:precision}), both of which show superior results using our approach. 
More specifically, \cref{fig:f1} shows that our strategy exhibits a consistently higher F1 score compared to other approaches, achieving the final result of $0.87$, which is better than other approaches, such as small-loss and feature-based~\cite{kim2021fine} that present a similar result of $0.75$. Another important comparison measure is precision. PASS shows very high precisions of more than $0.96$, while small-loss~\cite{li2020dividemix} and feature-based~\cite{kim2021fine} show much lower precision values around $0.72$. This empirical analysis suggests that our method is more efficacious at correctly identifying positive and negative samples from the training set than other competing approaches.

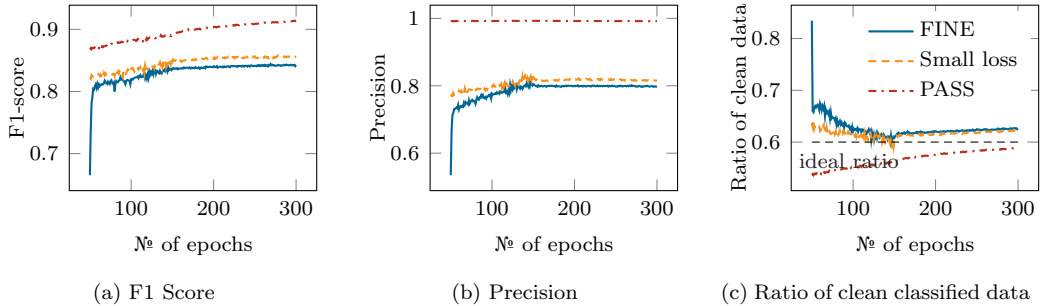
\begin{figure*}[t]
    \centering
    \begin{subfigure}[b]{0.3\textwidth}
        \centering
        \begin{tikzpicture}
            \pgfplotstableread[col sep=comma, header=true]{supp_cifar0.4_csv/f1.tex} \myTable
            \begin{axis}[
                height = 0.6\linewidth,
                width = 0.8\linewidth,
                xlabel={\textnumero~of epochs},
                xlabel style={font=\scriptsize},
                xticklabel style = {font=\scriptsize},
                ylabel={F1-score},
                ylabel style={font=\scriptsize, yshift=-0.5em},
                yticklabel style = {font=\scriptsize},
                scale only axis
            ]
                \addplot[mark=none, MidnightBlue, thick, solid] table[x={epochs}, y={FINE}]{\myTable};
                \addplot[mark=none, BurntOrange, thick, densely dashed] table[x={epochs}, y={SmallLoss}]{\myTable};
                \addplot[mark=none, BrickRed, solid, thick, dashdotted] table[x={epochs}, y={PASS}]{\myTable};
            \end{axis}
        \end{tikzpicture}
        \caption{F1 Score}
        \label{fig:0.4_f1}
    \end{subfigure}
    \hfill
    \begin{subfigure}[b]{0.3\textwidth}
        \centering
        \begin{tikzpicture}
            \pgfplotstableread[col sep=comma, header=true]{supp_cifar0.4_csv/precision.tex} \myTable
            \begin{axis}[
                height = 0.6\linewidth,
                width = 0.8\linewidth,
                xlabel={\textnumero~of epochs},
                xlabel style={font=\scriptsize},
                xticklabel style = {font=\scriptsize},
                ylabel={Precision},
                ylabel style={font=\scriptsize, yshift=-0.5em},
                yticklabel style = {font=\scriptsize},
                scale only axis
            ]
                \addplot[mark=none, MidnightBlue, thick, solid] table[x={epochs}, y={FINE}]{\myTable};
                \addplot[mark=none, BurntOrange, thick, densely dashed] table[x={epochs}, y={SmallLoss}]{\myTable};
                \addplot[mark=none, BrickRed, solid, thick, dashdotted] table[x={epochs}, y={PASS}]{\myTable};
            \end{axis}
        \end{tikzpicture}
        \caption{Precision}
        \label{fig:0.4_precision}
    \end{subfigure}
    \hfill
    \begin{subfigure}[b]{0.3\textwidth}
        \centering
        \begin{tikzpicture}
            \pgfplotstableread[col sep=comma, header=true]{supp_cifar0.4_csv/clean.tex} \myTable

            \begin{axis}[
                height = 0.6\linewidth,
                width = 0.8\linewidth,
                xlabel={\textnumero~of epochs},
                xlabel style={font=\scriptsize},
                xticklabel style = {font=\scriptsize},
                ylabel={Ratio of clean data},
                ylabel style={font=\scriptsize, yshift=-0.5em},
                yticklabel style = {font=\scriptsize},
                legend entries={FINE, Small loss, PASS},
                legend style={draw=none, font=\scriptsize},
                legend image post style={scale=1.},
                legend cell align={left},
                legend pos=north east,
                scale only axis
            ]
                \addplot[mark=none, MidnightBlue, thick, solid] table[x={epochs}, y={FINE}]{\myTable};
                \addplot[mark=none, BurntOrange, thick, densely dashed] table[x={epochs}, y={SmallLoss}]{\myTable};
                \addplot[mark=none, BrickRed, solid, thick, dashdotted] table[x={epochs}, y={PASS}]{\myTable};

                \addplot[mark=none, style={densely dashed}] coordinates {
                    (50, 0.6)
                    (300, 0.6)
                };
                \node[below, color=Black, rotate=0, ] at (95, 0.6) {\scriptsize{ideal ratio}};
            \end{axis}
        \end{tikzpicture}
        \caption{Ratio of clean classified data}
        
        \label{fig:0.4_clean}
    \end{subfigure}
    \caption{Graphs to compare the effectiveness of selecting clean or noisy samples, regarding three metrics: (a) F1-score, (b) precision, and (c) ratio of data classified as clean. The comparison is made between our PASS (dash-dot \textcolor{BrickRed}{red}), small-loss approach~\cite{jiang2018mentornet} (dashed \textcolor{BurntOrange}{yellow}),  and feature-based approach FINE~\cite{kim2021fine} (solid \textcolor{MidnightBlue}{blue}) (all on base model DivideMix~\cite{li2020dividemix}), implemented on the CIFAR-100 dataset\cite{krizhevsky2009learning} at \(0.4\) IDN noise rate, as described in~\cite{xia2020part}.}\label{fig:0.4_supp}
\end{figure*}

\paragraph{IDN settings at low and intermediate noise rates} (e.g., \(20\% \text{ and } 40\%\)) We further extended our empirical analysis to include other challenging IDN noise cases~\cite{xia2020part} at rates of $40\%$ and $20\%$, as shown in \cref{fig:0.4_supp,fig:0.2_supp} on CIFAR-100~\cite{krizhevsky2009learning} respectively. These plots compare our PASS (using DivideMix~\cite{li2020dividemix}) against the small-loss~\cite{li2020dividemix} and feature-based~\cite{kim2021fine} approaches by measuring the 
classification performance of clean samples based on (a) F1 score, (b) precision, and (c) ratio of data classified as clean. 

\begin{figure*}[t]
    \begin{subfigure}[b]{0.3\textwidth}
        \centering
        \begin{tikzpicture}
            \pgfplotstableread[col sep=comma, header=true]{supp_cifar0.2_csv/f1.tex} \myTable
            \begin{axis}[
                height = 0.6\linewidth,
                width = 0.8\linewidth,
                xlabel={\textnumero~of epochs},
                xlabel style={font=\scriptsize},
                xticklabel style = {font=\scriptsize},
                ylabel={F1-score},
                ylabel style={font=\scriptsize, yshift=-0.5em},
                yticklabel style = {font=\scriptsize},
                scale only axis
            ]
                \addplot[mark=none, MidnightBlue, thick, solid] table[x={epochs}, y={FINE}]{\myTable};
                \addplot[mark=none, BurntOrange, thick, densely dashed] table[x={epochs}, y={SmallLoss}]{\myTable};
                \addplot[mark=none, BrickRed, solid, thick, dashdotted] table[x={epochs}, y={PASS}]{\myTable};
            \end{axis}
        \end{tikzpicture}
        \caption{F1 Score}
        \label{fig:0.2_f1}
    \end{subfigure}
    \hfill
    \begin{subfigure}[b]{0.3\textwidth}
        \centering
        \begin{tikzpicture}
            \pgfplotstableread[col sep=comma, header=true]{supp_cifar0.2_csv/precision.tex} \myTable
            \begin{axis}[
                height = 0.6\linewidth,
                width = 0.8\linewidth,
                xlabel={\textnumero~of epochs},
                xlabel style={font=\scriptsize},
                xticklabel style = {font=\scriptsize},
                ylabel={Precision},
                ylabel style={font=\scriptsize, yshift=-0.5em},
                yticklabel style = {font=\scriptsize},
                legend entries={FINE, Small loss, PASS},
                legend style={draw=none, font=\scriptsize},
                legend image post style={scale=1.},
                legend cell align={left},
                legend pos=south east,
                scale only axis
            ]
                \addplot[mark=none, MidnightBlue, thick, solid] table[x={epochs}, y={FINE}]{\myTable};
                \addplot[mark=none, BurntOrange, thick, densely dashed] table[x={epochs}, y={SmallLoss}]{\myTable};
                \addplot[mark=none, BrickRed, solid, thick, dashdotted] table[x={epochs}, y={PASS}]{\myTable};
            \end{axis}
        \end{tikzpicture}
        \caption{Precision}
        \label{fig:0.2_precision}
    \end{subfigure}
    \hfill
    \begin{subfigure}[b]{0.3\textwidth}
        \centering
        \begin{tikzpicture}
            \pgfplotstableread[col sep=comma, header=true]{supp_cifar0.2_csv/clean.tex} \myTable
            \begin{axis}[
                height = 0.6\linewidth,
                width = 0.8\linewidth,
                xlabel={\textnumero~of epochs},
                xlabel style={font=\scriptsize},
                xticklabel style = {font=\scriptsize},
                ylabel={Ratio of clean data},
                ylabel style={font=\scriptsize, yshift=-0.5em},
                yticklabel style = {font=\scriptsize},
                scale only axis
            ]
                \addplot[mark=none, MidnightBlue, thick, solid] table[x={epochs}, y={FINE}]{\myTable};
                \addplot[mark=none, BurntOrange, thick, densely dashed] table[x={epochs}, y={SmallLoss}]{\myTable};
                \addplot[mark=none, BrickRed, solid, thick, dashdotted] table[x={epochs}, y={PASS}]{\myTable};

                \addplot[mark=none, style={densely dashed}] coordinates {
                    (50, 0.8)
                    (300, 0.8)
                };
                \node[above, color=Black, rotate=0, ] at (120, 0.8) {\scriptsize{ideal ratio}};
            \end{axis}
        \end{tikzpicture}
        \caption{Ratio of clean classified data}
        \label{fig:0.2_clean}
    \end{subfigure}
    \caption{Graphs to compare the effectiveness of selecting clean or noisy samples, regarding three metrics: (a) F1-score, (b) precision, and (c) ratio of data classified as clean. The comparison is made between our PASS (dash-dot \textcolor{BrickRed}{red}), small-loss approach~\cite{jiang2018mentornet} (dashed \textcolor{BurntOrange}{yellow}), and feature-based approach FINE~\cite{kim2021fine} (solid \textcolor{MidnightBlue}{blue}) (all on base model DivideMix~\cite{li2020dividemix}), implemented on the CIFAR-100 dataset\cite{krizhevsky2009learning} at \(0.2\) IDN noise rate, as described in~\cite{xia2020part}.}\label{fig:0.2_supp}
\end{figure*}

 From \cref{fig:0.4_supp,fig:0.2_supp}, it is clear that as training evolves, PASS gets closer to the ideal proportion of clean-label samples available for training than the small-loss~\cite{li2020dividemix} and feature-based~\cite{kim2021fine} approaches, suggesting that our approach
is more capable of identifying the correct proportion of noisy-label samples for the IDN noise. 
This proportion alone does not imply accuracy. Therefore, we also provide graphs with F1 and precision scores, which help to highlight the advantages of using our peer agreement for sample selection. 
More specifically, \cref{fig:0.4_supp,fig:0.2_supp} show that our strategy exhibits a consistently superior F1 score compared to other approaches for noise rates $40\%$ (\cref{fig:0.4_f1}) and $20\%$ (\cref{fig:0.2_f1}). PASS achieves a final result of $0.92$, which directly reflects the improvement in the performance of PASS when compared to small-loss~\cite{li2020dividemix} and feature-based~\cite{kim2021fine} with similar results of $0.8-0.85$ at noise rate $40\%$. Whilst feature-based~\cite{kim2021fine} and PASS are very competitive in F1 score for noise rate $20\%$ with a value around $0.94$, small-loss~\cite{li2020dividemix} stays around $0.89$.
PASS shows an outstanding precision higher than $0.98$, while small-loss~\cite{li2020dividemix} and feature-based~\cite{kim2021fine} show much smaller precision values of around $0.8$ for noise rate $40\%$ (\cref{fig:0.4_precision}). Moreover, all methods are very competitive in precision at a low noise rate of $20\%$ (\cref{fig:0.2_precision}). Our empirical analysis shows that our method outperforms other competing approaches in correctly identifying positive and negative samples from the training set across all noise levels.

\subsubsection{Empirical Insights on PASS using Clothing1M}\label{subsec:empirical_subsec_clothing}

\begin{figure*}[t!]
    \centering
    \begin{subfigure}[b]{0.3\textwidth}
        \centering
        \begin{tikzpicture}
            \pgfplotstableread[col sep=comma, header=true]{clothing1m_csv/f1.tex} \myTable
            \begin{axis}[
                height = 0.6\linewidth,
                width = 0.8\linewidth,
                xlabel={\textnumero~of epochs},
                xlabel style={font=\scriptsize},
                xticklabel style = {font=\scriptsize},
                ylabel={F1-score},
                ylabel style={font=\scriptsize, yshift=-0.5em},
                yticklabel style = {font=\scriptsize},
                legend entries={AugDesc, AugDesc-PASS},
                legend style={draw=none, font=\scriptsize},
                legend image post style={scale=0.5},
                legend cell align={left},
                legend pos=south east,
                scale only axis
            ]
                \addplot[mark=none, BurntOrange, thick, densely dashed] table[x={epochs}, y={SmallLoss}]{\myTable};
                \addplot[mark=none, MidnightBlue, solid, thick, solid] table[x={epochs}, y={PASS}]{\myTable};
            \end{axis}
        \end{tikzpicture}
        \caption{F1 Score}
        \label{fig:cloth_f1}
    \end{subfigure}
    \hfill
    \begin{subfigure}[b]{0.3\textwidth}
        \centering

        \begin{tikzpicture}
            \pgfplotstableread[col sep=comma, header=true]{clothing1m_csv/precision_.tex} \myTable

            \begin{axis}[
                height = 0.6\linewidth,
                width = 0.8\linewidth,
                xlabel={\textnumero~of epochs},
                xlabel style={font=\scriptsize},
                xticklabel style = {font=\scriptsize},
                ylabel={Precision},
                ylabel style={font=\scriptsize, yshift=-0.5em},
                yticklabel style = {font=\scriptsize},
                scale only axis
            ]
                \addplot[mark=none, BurntOrange, thick, densely dashed] table[x={epochs}, y={SmallLoss}]{\myTable};
                \addplot[mark=none, MidnightBlue, solid, thick, solid] table[x={epochs}, y={PASS}]{\myTable};
            \end{axis}
        \end{tikzpicture}
        \caption{Precision}
        \label{fig:cloth_precision}
    \end{subfigure}
    \hfill
    \begin{subfigure}[b]{0.3\textwidth}
        \centering

        \begin{tikzpicture}
            \pgfplotstableread[col sep=comma, header=true]{clothing1m_csv/clean.tex} \myTable
            \begin{axis}[
                height = 0.6\linewidth,
                width = 0.8\linewidth,
                xlabel={\textnumero~of epochs},
                xlabel style={font=\scriptsize},
                xticklabel style = {font=\scriptsize},
                ylabel={Ratio of clean data},
                ylabel style={font=\scriptsize, yshift=-0.5em},
                yticklabel style = {font=\scriptsize},
                scale only axis
            ]
                \addplot[mark=none, BurntOrange, thick, densely dashed] table[x={epochs}, y={SmallLoss}]{\myTable};
                \addplot[mark=none, MidnightBlue, solid, thick, solid] table[x={epochs}, y={PASS}]{\myTable};
            \end{axis}
        \end{tikzpicture}
        \caption{Ratio of clean classified data}
        \label{fig:cloth_ratio}
    \end{subfigure}
    \caption{Graphs to compare the effectiveness of selecting clean or noisy samples, regarding three metrics: (a) F1-score, (b) precision, and (c) ratio of data classified as clean.
    The comparison is made between our approach AugDesc-PASS (solid \textcolor{MidnightBlue}{blue}) with baseline AugDesc approach~\cite{nishi2021augmentation} (dashed \textcolor{BurntOrange}{yellow}) on Clothing1M. We have used DM-AugDesc-WS-WAW version of training as mentioned in AugDesc~\cite{nishi2021augmentation}.}\label{fig:clothing_supp}
\end{figure*}
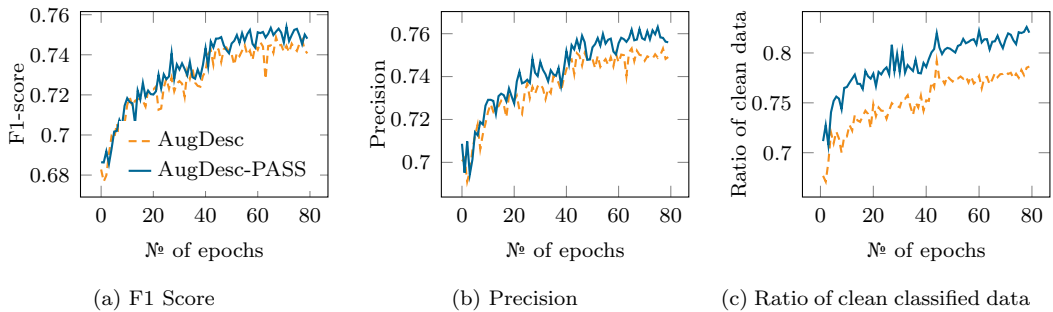

Although Clothing1M~\cite{xiao2015learning} offers a clean validation set, we did not incorporate it into our training process. However, we used this clean validation set to assess and compare the effectiveness of PASS and baseline AugDesc~\cite{nishi2021augmentation}. For AugDesc training with and without PASS, we have used the \emph{DM-AugDesc-WS-WAW} version of training, as mentioned in AugDesc~\cite{nishi2021augmentation}. As mentioned in \cref{table:clothing1M},  
our results are competitive with the existing model. Although both baseline methods are competitive, PASS is still capable of outperforming based on: (\subref{fig:cloth_f1}) F1, (\subref{fig:cloth_precision}) precision, and (\subref{fig:cloth_ratio}) the ratio of clean data in \cref{fig:clothing_supp}.

\section{Conclusion}
\label{sec:conclusion}
This article proposed a new peer-agreement-based sample selection technique, PASS, for noisy-label learning
to improve the performance of robust classifiers.
We show that PASS can be easily integrated into SOTA noisy-label learning methods~\cite{li2020dividemix, Garg_2023_WACV, xu2021faster, feng2021ssr, zheltonozhskii2022contrast, nishi2021augmentation}
to improve their classification accuracy results on several noisy-label learning benchmarks, including CIFAR-100~\cite{krizhevsky2009learning}, Red mini-ImageNet from CNWL~\cite{xu2021faster}, Animal-10N~\cite{song2019selfie}, CIFAR-N~\cite{wei2022learning}, Clothing1M~\cite{xiao2015learning}, mini-Webvision~\cite{li2017webvision}, and Imagenet~\cite{krizhevsky2012imagenet}. It consistently outperforms existing methods in most cases. 
Our proposed approach has the potential to create a positive societal impact by mitigating biases in resolving noisy-labelled data. 
The slight increase in training time, between \((2\%)\) and \((10\%)\) as detailed in~\cref{sec:computational_time}, is a small investment for the gains in accuracy and reliability of the model. 
Furthermore, in addition to the gains listed above, our strategic design choice enables a richer, and more nuanced understanding of the data. 
Looking ahead, we plan to refine and enhance PASS's efficiency through methods such as dimensionality reduction and early stopping, alongside the adoption of mixed precision training. The integration of these techniques will not only streamline PASS's performance, but also significantly broaden its applicability and effectiveness in diverse scenarios, solidifying its position as a state-of-the-art tool in the field.

\bibliographystyle{elsarticle-num} 
\bibliography{ref}




\end{document}